\documentclass[10pt,journal]{IEEEtran}

\usepackage[english]{babel}
\usepackage[utf8x]{inputenc}
\usepackage[T1]{fontenc}
\usepackage{makecell}

\usepackage{amsmath,amssymb,mathrsfs}
\usepackage{graphicx}
\usepackage[colorinlistoftodos]{todonotes}
\usepackage[colorlinks=true, allcolors=blue]{hyperref}
\usepackage{multirow}
\usepackage[lined,linesnumbered,ruled]{algorithm2e}
\usepackage[noadjust]{cite}
\usepackage[colorinlistoftodos]{todonotes}

\usepackage{xfrac}

\newcommand{\cp}[1]{\ifmmode {\mathcal{#1}}\else ${\mathcal{#1}}$\fi}
	
\newcommand{\bA}{\boldsymbol{A}}

\newcommand{\bI}{\boldsymbol{I}}

\newcommand{\bN}{\boldsymbol{N}}

\newcommand{\bU}{\boldsymbol{U}}
\newcommand{\bV}{\boldsymbol{V}}
\newcommand{\bW}{\boldsymbol{W}}
\newcommand{\bX}{\boldsymbol{X}}
\newcommand{\bY}{\boldsymbol{Y}}
\newcommand{\bzero}{\boldsymbol{0}}

\newcommand{\bz}{\boldsymbol{z}}

\newcommand{\bOmega}{\boldsymbol{\Omega}}

\usepackage{color}  % To include comments in color

\def\cblue{}

\definecolor{darkgreen}{rgb}{0.1, 0.80, 0.32}
\title{A Fast Multiscale Spatial Regularization \\for Sparse Hyperspectral Unmixing}

\author{Ricardo~Augusto~Borsoi, Tales Imbiriba,~\IEEEmembership{Member,~IEEE,} \\Jos\'e~Carlos~Moreira~Bermudez,~~\IEEEmembership{Senior~Member,~IEEE}, C\'edric Richard,~~\IEEEmembership{Senior~Member,~IEEE}
\thanks{This work has been supported by the National Council for Scientific and Technological Development (CNPq).}
\thanks{R.A. Borsoi, T. Imbiriba and J.C.M. Bermudez are with the Department
of Electrical Engineering, Federal University of Santa Catarina, Florian\'opolis, SC, Brazil. e-mail: raborsoi@ucs.br; talesim@gmail.com; j.bermudez@ieee.org. C. Richard is with the Universit\'e C\^ote d'Azur, Nice, France (e-mail: cedric.richard@unice.fr), Lagrange Laboratory (CNRS, OCA).}% <-this % stops a space
\thanks{Manuscript received Month day, year; revised Month day, year.}}

\begin{document}
\maketitle

\begin{abstract}
Sparse hyperspectral unmixing from large spectral libraries has been considered to circumvent limitations of endmember extraction algorithms in many applications.
This strategy often leads to ill-posed inverse problems, which can greatly benefit from spatial regularization strategies. However, existing spatial regularization strategies lead to large-scale nonsmooth optimization problems. Thus, efficiently introducing spatial context in the unmixing problem remains a challenge, and a necessity for many real world applications.
In this paper, a novel multiscale spatial regularization approach for sparse unmixing is proposed. The method uses a signal-adaptive spatial multiscale decomposition based on segmentation and over-segmentation algorithms to decompose the unmixing problem into two simpler problems, one in an approximation image domain and another in the original domain.
Simulation results using both synthetic and real data indicate that the proposed method outperforms state-of-the-art Total Variation-based algorithms with a computation time comparable to that of their unregularized counterparts.

% Your abstract.
% ==> say in results that proposed is better specially at low SNR in synthetic data, and that in real data is similar because Cuprite image has high SNR and cite Iordache.\\
% ==> colocar colorbar encima ou embaixo das figuras\\
% ==> DC2 generated using a dirichlet distribution over a Matern Gaussian Field, giving a correlated with slight variability and steep transitions\\
% ==> our method is less prone to oversmoothing than total variation, hence deals better with fine spatial variability of abundance maps
\end{abstract}


\section{Introduction}

Hyperspectral unmixing is at the core of many remote sensing and earth observation applications~\cite{Bioucas2012}. The limited spatial resolution of hyperspectral devices often mixes the spectral contributions of different pure materials, named \emph{endmembers}, in the scene~\cite{Keshava:2002p5667}. The mixing process conceals crucial information relating the endmembers and their spatial disposition. Hyperspectral unmixing (HU) aims to solve this problem by separating the hyperspectral image (HI) into a collection of endmembers and their fractional \emph{abundances}~\cite{Bioucas2012}.

Notwithstanding the relevance of more complex aspects in modeling the mixing process such as nonlinearity~\cite{Imbiriba2016_tip} and spectral variability~\cite{imbiriba2018GLMM}, most unmixing methods use a simple linear mixing model (LMM). The LMM considers an observed reflectance vector (a pixel) to be a convex combination of endmembers~\cite{Keshava:2002p5667}. The combination coefficients are termed abundances, as each one can be interpreted as the proportion of the pixel spectrum contributed by the corresponding endmember~\cite{Keshava:2002p5667}. The LMM leads \cblue{to} computationally efficient solutions and yields high quality unmixing results for several applications.

The problems of endmember extraction and unmixing are interrelated and addressing them jointly is not always trivial. Most endmember extraction algorithms rely on the existence of pure pixels or on the data not being heavily mixed~\cite{Ma-2014-ID323,Imbiriba2016_tip}.

%\cblue{[OBS: The reference is not on endmember extraction. We should use a more specific reference here.]}
%
An interesting strategy to circumvent such issues is to model the observed pixel as a linear combination of a large library of endmembers estimated \emph{a priori}~\cite{Bioucas2012}.
In this case, the number of endmembers in a given scene is usually much smaller than the size of the spectral library. Hence, the unmixing problem becomes a sparse regression problem that consists of finding a small subset of the library endmembers which best represent all the pixels in the image. This problem is often efficiently solved through the use of sparsity promoting regularizations, resulting in the so-called sparse unmixing techniques~\cite{bioucas2010SUnSAL}.

Despite the success of standard sparse unmixing methods~\cite{bioucas2010SUnSAL}, the use of large spectral libraries leads to the unmixing problem being ill-posed, which makes the solution very sensitive to noise.
Regularization techniques have been shown to significantly improve the performance of both non-sparse~\cite{lu2014doubleConstrainedNMF,imbiriba2018ULTRA} and sparse unmixing methods~\cite{iordache2012sparseUnmixingTV} by exploiting the correlation between different pixels in the HI. % shi2014surveySpatialRegUnmixing
% \cred{[...tensor... or employing clustering strategies... cite this \cite{lu2014doubleConstrainedNMF}]}
%
The \emph{Total Variation} (TV) regularization, for instance, promotes solutions that are spatially piecewise homogeneous without compromising sharp discontinuities between neighboring pixels~\cite{iordache2012sparseUnmixingTV}.

% Despite the success of spatial regularization techniques in sparse unmixing, their adoption has come at the expense of a massive increase in computational cost.
Most effective spatial regularization techniques, however, require a massive increase in computational cost.
%
% For instance, the TV regularization considered in~\cite{iordache2012sparseUnmixingTV} to tackle the sparse unmixing problem leads to a large non-smooth convex optimization problem, which needs to be solved using variable splitting techniques. More recently, regularization strategies exploiting nonlocal redundancy in images were also considered, leading to even larger optimization problems~\cite{wang2017centralizedNonlocalSparseUnmixing}.
For instance, TV~\cite{iordache2012sparseUnmixingTV} leads to a large non-smooth convex optimization problem, which needs to be solved using variable splitting techniques. More recently, regularization strategies exploiting nonlocal redundancy in images were also considered, leading to even larger optimization problems~\cite{wang2017centralizedNonlocalSparseUnmixing}.
This is incompatible with recent demands to timely process the vast amounts of remotely sensed data required by many real world applications~\cite{chi2016BigDataRemoteSensing}. Such demands recently sparked significant interest on efficient unmixing strategies with online processing capability~\cite{bernabe2018GPUunmixing}.
This evidences the need for fast low complexity unmixing strategies that yield state of the art performance.

% =====

This paper introduces a novel multiscale spatial regularization approach for sparse unmixing. We propose a fast \emph{Multiscale sparse Unmixing Algorithm} (MUA) that promotes piecewise homogeneous abundances without compromising sharp discontinuities among neighboring pixels. 
The proposed method uses a signal-adaptive spatial multiscale decomposition of the linear mixture model. The unmixing problem is decomposed into two different problems in distinct domains: one in an approximation scale representation constructed using segmentation or over-segmentation algorithms, and another in the original image domain.
Spatial contextual information of fractional abundances is initially obtained by solving an unregularized sparse unmixing problem in the approximation scale. 
This information is then mapped back to the original image domain by means of an appropriately defined conjugate transformation of the multiscale decomposition. The spatial contextual information is then enforced on the solution of the original unmixing problem through a novel and computationally efficient regularization penalty.
%
%and is then used to efficiently introduce context-aware spatial regularity into the solution of the original unmixing problem in the form of a computationally efficient regularization penalty.
%
Simulation results using both synthetic and real data indicate that the proposed method outperforms TV-regularized solutions~\cite{iordache2012sparseUnmixingTV}, \cblue{while} requiring a computational time comparable to that of the unregularized algorithm~\cite{bioucas2010SUnSAL}.

The paper is organized as follows. In Section~\ref{sec:unmixing_model}, we briefly introduce the sparse unmixing problem and present the proposed multiscale formulation.
Simulation results using synthetic and real data are presented in Section~\ref{sec:results}.
Section~\ref{sec:conclusions} presents the conclusions.

% \clearpage

\section{Sparse linear unmixing with a multiscale spatial regularization}
\label{sec:unmixing_model}

Let $\bY\in\mathbb{R}^{L\times N}$ denote the observed hyperspectral image with $L$ bands and $N$ pixels, and $\bA\in\mathbb{R}^{L\times P}$ denote a spectral library having $P$ spectral signatures.
Instead of extracting the endmembers directly from the HI $\bY$, sparse linear unmixing attempts to find an optimal subset of samples from the spectral library $\bA$ that best represents all the mixed pixels in the image, namely,
\begin{align} \label{eq:sp_model_eq}
	\bY = \bA \bX + \bN 
    \,\text{,}
\end{align}
where $\bX\in\mathbb{R}^{P\times N}$ is the fractional abundance matrix, each column of which determines the composition of one image pixel as a linear combination of spectral samples from $\bA$, and $\bN\in\mathbb{R}^{L\times N}$ denotes the joint contribution of modeling errors and noise.
The fractional abundance matrix $\bX$ is frequently subject to physical constraints  imposed to the model, such as the non-negativity (i.e. $x_{i,j}\ge 0$, $\forall i,j$, denoted by $\bX\geq\bzero$) and the sum-to-one constraints (i.e. $\mathbf{1}^\top\bX=\mathbf{1}^\top$). 
Since only few of the spectral signatures of $\bA$ are likely to contribute to the observed spectra of each pixel, the matrix $\bX$ is usually sparse.
%
%
% \cred{A common approach to solve the unmixing problem is to represent it as a sparse regression problem, using spatial regularization to overcome the ill-posedness of the inverse problem~\cite{iordache2012sparseUnmixingTV}. These techniques, however, are computationally very expensive. In this section, we propose a spatial regularization procedure based on a multi-scale transformation which introduces spatial regularity into the abundance maps at a very low computational cost.}
% \cred{A common approach to solve the unmixing problem is to represent it as a spatially regularized sparse regression problem~\cite{iordache2012sparseUnmixingTV}. These techniques, however, are computationally very expensive. In this section, we propose a regularization procedure based on a multi-scale transformation which introduces spatial regularity into the abundance maps at a very low computational cost.}
A common approach to solve the unmixing problem is to represent it as a spatially regularized sparse regression problem~\cite{iordache2012sparseUnmixingTV}. These techniques, however, are computationally very expensive. In this section, we propose a multiscale regularization procedure which introduces spatial regularity into the abundance maps at a very low computational cost.

The proposed spatially regularized unmixing scheme consists of two steps. First, we transform the original image from the original domain ($\mathcal{D}$) to an approximation (coarse) scale ($\mathcal{C}$) to extract the most relevant inter-pixel contextual information. Then, pixels at the coarse scale are unmixed independently from each other. Next, we apply a conjugate transformation to the abundance estimates obtained at the coarse scale to convert the coarse estimate back to the original image domain. This procedure yields an accurate estimate of the low-level image structures, which is then used to regularize the unmixing process applied to the original image to promote the spatial dependency between neighboring pixels.

% The proposed spatially regularized unmixing scheme consists of two steps. First, we perform a spatial transformation in the input image to convert it to an approximation/coarse scale, where the pixels can be unmixed independently from each other. Afterwards, a conjugate transformation is applied to the generated abundances, converting the coarse estimative back to the image domain. This generates an accurate estimate of the low-level image structures, which is then used to regularize the unmixing of the original image, enforcing the spatial correlation between neighboring pixels.

Consider a linear operator~$\bW \in \mathbb{R}^{N\times K}$, $K < N$ that implements a spatial transformation of both the HI and the abundance map to the approximation domain. Then,
\begin{equation}
\begin{split}
	\bY_{\!\mathcal{C}} = \bY\bW \,;\qquad \bX_{\!\mathcal{C}} = \bX\bW
    \,\text{,}
\end{split}
\end{equation}
where $\bY_{\!\mathcal{C}} \in \mathbb{R}^{L\times K}$ and $\bX_{\!\mathcal{C}} \in \mathbb{R}^{P\times K}$ are the coarse approximations of the original image $\bY$ and of the abundance matrix $\bX$, respectively.
A possible choice for $\bW$ might be a wavelet transform employing the first $K$ approximation scales of the wavelet decomposition of $\bY$. However, the wavelet transform is feature-agnostic. It does not distinguish between pixels in perceptually different image regions. Its application may result in blurred image edges.
Instead, we shall consider a signal-dependent transformation, that is, $\bW\equiv\bW(\bY)$, which groups pixels into perceptually meaningful regions (not necessarily uniform), preserving image contours and leading to sharp transitions.

Multiplying~\eqref{eq:sp_model_eq} by $\bW$ from the right, the unmixing problem can be re-cast into the approximation domain. The resulting unmixing problem is as follows:
\begin{equation} \label{eq:prob_appr_scale}
\begin{split}
    \widehat{\!\bX}_{\!\mathcal{C}} {}={} & 
    \mathop{\arg\min}_{\bX_{\!\mathcal{C}}\geq0} \,\,\, 
    \resizebox{0.25cm}{!}{${\displaystyle\frac{1}{2}}$}
%     \frac{1}{2}
    \|\bY_{\!\mathcal{C}} -\bA\bX_{\!\mathcal{C}} \|_{F}^2 + \lambda_{\mathcal{C}} \|\bX_{\!\mathcal{C}}\|_{1,1}
    \,.
\end{split}
\end{equation}
% \begin{equation} \label{eq:prob_appr_scale}
% \begin{split}
%     \widehat{\!\bX}_{\!\mathcal{C}} {}={} & 
%     \mathop{\arg\min}_{\bX_{\!\mathcal{C}}\geq0} \,\,\, \|\bY_{\!\mathcal{C}} -\bA\bX_{\!\mathcal{C}} \|_{F}^2 + 2\,\lambda_{\mathcal{C}} \|\bX_{\!\mathcal{C}}\|_{1,1}
%     \,.
% \end{split}
% \end{equation}

We shall now use $\widehat{\!\bX}_{\!\mathcal{C}}$ to regularize the original unmixing problem. To this end, we define a conjugate transform $\bW^{\ast}\in \mathbb{R}^{K\times N}$ that converts images from the ap\-prox\-i\-ma\-tion domain $\mathcal{C}$ back to the original image domain $\mathcal{D}$:
\begin{equation} \label{eq:conjugate_mscale_transf}
\begin{split}
	 \widehat{\!\bX}_{\!\mathcal{D}} = \widehat{\!\bX}_{\!\mathcal{C}}\bW^{\ast}
    \,\text{,}
\end{split}
\end{equation}
where $\widehat{\!\bX}_{\!\mathcal{D}}\in\mathbb{R}^{P\times N}$ is the low-resolution approximation of the abundances in the original image domain, which captures correlations between neighboring pixels. Note that transformation $\bW$ is generally not invertible, that is, $\bW\bW^{\ast}\neq\bI$.

% By the definition of the desired properties of $\bW$,   abundance vectors in $\widehat{\!\bX}_{\!\mathcal{D}}$ is highly correlated with pixels in similar neighboring regions. Therefore, by enforcing the estimated abundances in $\bX$ to be similar to the approximation image $\widehat{\!\bX}_{\!\mathcal{D}}$, we are forcing it to be similar to its neighbors.
%
Finally, we use the coarse abundance matrix $\widehat{\!\bX}_{\!\mathcal{D}}$ to regularize a sparse unmixing problem in the original image domain, where $\,\widehat{\!\bX}$ is obtained as the solution to the following optimization problem:
\begin{equation} \label{eq:prob_orig_scale_reg}
\begin{split}
    \mathop{\min}_{\bX\geq0}\,\,\,
 	\resizebox{0.25cm}{!}{${\displaystyle\frac{1}{2}}$}
%     \frac{1}{2}
    \|\bY -\bA\bX \|_{F}^2 + \lambda \|\bX\|_{1,1} 
	+ \resizebox{0.28cm}{!}{${\displaystyle\frac{\beta}{2}}$}
%     + \frac{\beta}{2}
    \|\,\widehat{\!\bX}_{\!\mathcal{D}} -\bX\|_{F}^2
    \,,
\end{split}
\end{equation}
% \begin{equation} \label{eq:prob_orig_scale_reg}
% \begin{split}
%     \mathop{\min}_{\bX\geq0}\,\,\,\|\bY -\bA\bX \|_{F}^2 + 2\,\lambda \|\bX\|_{1,1} 
%     + \beta \|\,\widehat{\!\bX}_{\!\mathcal{D}} -\bX\|_{F}^2
%     \,,
% \end{split}
% \end{equation}
where $\beta$ is a regularization parameter. \cblue{This formulation requires no explicit consideration of dependencies between pairs of pixels as required by TV. This leads to a simpler optimization problem, reducing both the computational complexity and the convergence time, as will be verified in Section~\ref{sec:results}.}

%Note that, unlike TV or \cblue{Tikhonov regularization}, the proposed method does not require to explicitly \cblue{consider dependencies} between pairs of pixels. This results in a computationally efficient procedure.

Note that both optimization problems~\eqref{eq:prob_appr_scale} and~\eqref{eq:prob_orig_scale_reg} are particular cases of the following problem
\begin{equation} \label{eq:base_admm}
	\min_{\bz} \,\,\, f_1(\bz) + f_2(\bz)
    \,,
\end{equation}
where $f_1,f_2:\mathbb{R}^n\to\mathbb{R}_+\cup\{\infty\}$ are closed, proper and convex functions.
For instance, problem~\eqref{eq:prob_orig_scale_reg} can be written in the equivalent form~\eqref{eq:base_admm} by selecting functions $f_1$ and $f_2$ as
\begin{align}
\begin{split}
	f_1 {}\equiv{} & 
%     \frac{1}{2}
    \resizebox{0.25cm}{!}{${\displaystyle \frac{1}{2}}$}
    \|\bY -\bA\bX \|_{F}^2 
%     + \frac{\beta}{2}
    + \resizebox{0.28cm}{!}{${\displaystyle \frac{\beta}{2}}$}
    \|\,\widehat{\!\bX}_{\!\mathcal{D}} -\bX\|_{F}^2 \\
	f_2 {}\equiv{} &  \lambda \|\bX\|_{1,1} + \iota_+(\bX) 
    \,,
\end{split}
\end{align}
% \begin{align}
% \begin{split}
% 	f_1 {}\equiv{} & \|\bY -\bA\bX \|_{F}^2 + \beta\|\,\widehat{\!\bX}_{\!\mathcal{D}} -\bX\|_{F}^2 \\
% 	f_2 {}\equiv{} &  2\,\lambda \|\bX\|_{1,1} + \iota_+(\bX) 
%     \,,
% \end{split}
% \end{align}
where $\iota_+(\cdot)$ is the indicator function of the set $\mathbb{R}^{P \times N}_+$, that is, $\iota_+(\bX)=0$ if $\bX\geq0$ and $\iota_+(\bX)=\infty$ otherwise.

The Alternating Direction Method of Multipliers (ADMM) method decomposes a problem in the form~\eqref{eq:base_admm} into a sequence of simpler problems, which can be solved efficiently~\cite{eckstein1992douglasSplittingADMM}.
The ADMM method can then be used to solve~\eqref{eq:prob_orig_scale_reg}, with the resulting procedure detailed in Algorithm~\ref{alg:ADMM} \cite{eckstein1992douglasSplittingADMM,bioucas2010SUnSAL}, where $\text{soft}$ denotes the component-wise soft thresholding operator $\text{soft}(y,\tau)= \text{sign}(y)\max\{|y|-\tau, 0 \}$. %~\cite{chen2001atomicBasisPursuitSIAM}.
Note that problem~\eqref{eq:prob_appr_scale} can be solved in the same way by setting $\beta=0$ and substituting $\bY\equiv\bY_{\mathcal{C}}$, $\bX\equiv\bX_{\mathcal{C}}$, and $\lambda\equiv\lambda_{\mathcal{C}}$ in Algorithm~\ref{alg:ADMM}.
The global algorithm of the proposed method, called \emph{Multiscale sparse Unmixing Algorithm} (MUA), is displayed in Algorithm~\ref{alg:proposed_alg}.

\begin{algorithm} [bth]
% \small
% \scriptsize
\footnotesize
\SetKwInOut{Input}{Input}
\SetKwInOut{Output}{Output}
\caption{ADMM method for solving \eqref{eq:prob_orig_scale_reg}~\label{alg:ADMM}}
\Input{$\bY$, $\bA$, parameters $\lambda$, $\beta$, and $\mu>0$ and matrices $\bU_0,\bV_0\in\mathbb{R}^{P\times N}$.}
\Output{The estimated abundance matrix $\widehat{\!\bX}$.}
Set $i=0$ \;
\While{stopping criterion is not satisfied}{
$\bOmega = \bA^\top\bY + \mu(\bU_i + \bV_i) + \beta \, \widehat{\!\bX}_{\!\mathcal{D}}$ \;
$\bX_{i+1} = \big(\bA^\top\bA + (\mu+\beta)\bI \big)^{-1} \bOmega$ \;
$\bU_{i+1} = \max\{\mathbf{0},\,\text{soft}(\bX_{i+1}-\bV_i,\,\lambda/\mu)\}$ \;
$\bV_{i+1} = \bV_i - (\bX_{i+1}-\bU_{i+1})$ \;
$i=i+1$ \;
}
\KwRet $\widehat{\!\bX}=\bX_{i+1}$\;
\end{algorithm}

\vspace{-0.5cm}
\begin{algorithm} [bth]
% \small
\footnotesize
\SetKwInOut{Input}{Input}
\SetKwInOut{Output}{Output}
\caption{MUA~\label{alg:proposed_alg}}
\Input{$\bY$, $\bA$, $\bW$, parameters $\lambda_{\mathcal{C}}$, $\lambda$, and $\beta$.}
\Output{The estimated abundance matrix $\widehat{\!\bX}$.}
Compute $\bY_{\mathcal{C}}=\bY\bW$ \; 
% Compute the superpixel decomposition, $\bY_{\!\mathcal{C}}$ of the HI $\bY$ using the SLIC algorithm~\cite{achanta2012slicPAMI}\; 
% Find $\widehat{\!\bX}_{\!\mathcal{C}}$ by solving~\eqref{eq:prob_appr_scale} using Algorithm~\ref{alg:ADMM}\;
\cblue{Find $\widehat{\!\bX}_{\!\mathcal{C}}$ by solving~\eqref{eq:prob_appr_scale} using Algorithm~\ref{alg:ADMM} with $\beta=0$, $\lambda\equiv\lambda_{\mathcal{C}}$, $\bY\equiv\bY_{\mathcal{C}}$ and $\bX\equiv\bX_{\mathcal{C}}$}\;
Compute $\widehat{\!\bX}_{\!\mathcal{D}}$ using~\eqref{eq:conjugate_mscale_transf}\;
Find $\widehat{\!\bX}$ by solving~\eqref{eq:prob_orig_scale_reg} using Algorithm~\ref{alg:ADMM}\;
\KwRet $\widehat{\!\bX}$\;
\end{algorithm}

%\vspace{-2cm}

\subsection{Designing the multiscale transformation}

An appropriate choice of transformation $\bW$ is of paramount importance for the proposed method to achieve a good reconstruction accuracy. The objectives of this transform can be summarized as 1) grouping image pixels that are spatially adjacent and semantically similar, that is, that belong to homogeneous regions, and 2) preserving image contours by not grouping pixels that belong to different image structures or features. Additionally, it must be computationally efficient.

Techniques such as the K-means have been explored for introducing regularity into the solution of inverse problems~\cite{lu2014doubleConstrainedNMF}. However, K-means fails to effectively explore local spatial regularity of the image, which is an important contextual information of HIs. Moreover, spectrally similar pixels might share different abundance attributes. Hence, spectral-only methods such as the K-means tend to group pixels that are semantically distinct, especially in noisy scenarios. Therefore, both spatial and spectral information should be explored to obtain good results.

% \cblue{Techniques such as the K-means have been explored for introducing regularity into the solution of inverse problems. However, K-means fails} to explore spatial regularity of the image, which is an important contextual \cblue{information of HIs. Moreover, spectrally similar pixels might share different abundance attributes. Hence,} spectral-only methods such as the K-means \cblue{tend to group pixels that} are semantically distinct, \cblue{especially in noisy scenarios. Therefore, both spatial and spectral information should be explored} to obtain good results.

To explore spatial information while grouping semantically similar pixels accounting for image discontinuities, we propose to construct~$\bW$ using image  segmentation or over-segmentation algorithms~\cite{veganzones2014hyperspectralSegmentationBPT,achanta2012slicPAMI}.
Image segmentation methods decompose the observed image into a set of contiguous homogeneous regions with contextually similar spatial information, typically consisting of objects which are separated by image borders~\cite{veganzones2014hyperspectralSegmentationBPT}.
Image segmentation often creates groups of pixels of heterogeneous sizes, corresponding to both small and large objects in the same image. Although this allows one to represent large regions with homogeneous abundance characteristics without compromising smaller objects, it can lead to grouping pixels that share different abundance characteristics (even if spectrally similar).
As an alternative to circumvent this issue, we also explore over-segmentation techniques, which attempt to divide the observed image into a larger number of regions with relatively homogeneous sizes~\cite{achanta2012slicPAMI}. Although over-segmentation methods partition large objects into many smaller segments, they provide an increased ability to adequately represent image borders and reduce the chance of grouping pixels with different contextual information. Superpixel algorithms are a popular and efficient technique for image over-segmentation~\cite{achanta2012slicPAMI}.
% tight adherence to image contours

\begin{figure}[bth]
\centering
\begin{minipage}[t]{.3\linewidth}
  \centering
  \centerline{\includegraphics[width=\linewidth,trim={0cm 3cm 0cm 0},clip]{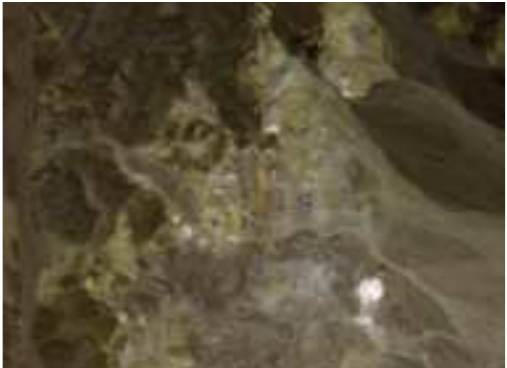}}
  \centerline{HS Image}\medskip
\end{minipage}
% \hfill
\hspace{0.1cm}
\begin{minipage}[t]{0.3\linewidth}
  \centering
  \centerline{\includegraphics[width=\linewidth,trim={0 3cm 0 0},clip]{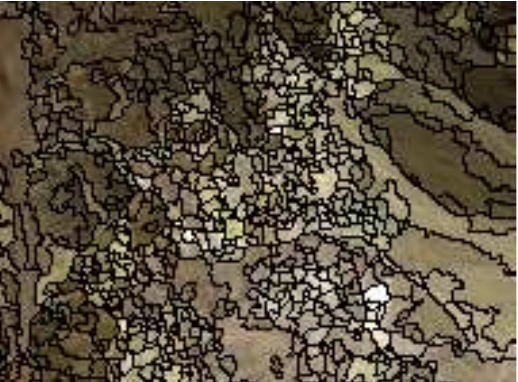}}
  \centerline{Segmentation}\medskip
\end{minipage}
% \hfill
\hspace{0.1cm}
\begin{minipage}[t]{0.3\linewidth}
  \centering
  \centerline{\includegraphics[width=\linewidth,trim={0 3cm 0 0},clip]{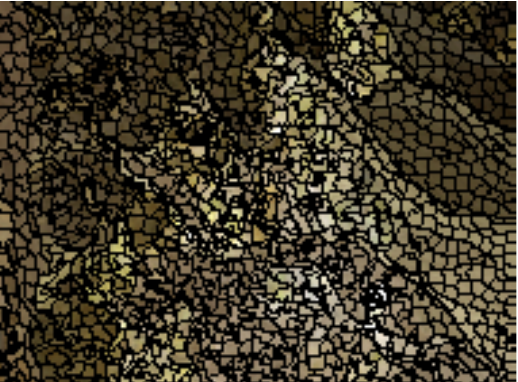}}
  \centerline{Over-Segmentation}\medskip
\end{minipage}
\vspace{-0.2cm}
\caption{Coarse-scale decomposition of a section of the Cuprite image for bands 50, 80 and 100 using the segmentation algorithm in~\cite{veganzones2014hyperspectralSegmentationBPT} and the over-segmentation algorithm in~\cite{achanta2012slicPAMI}, with 950 and 2000 segments, respectively.}
%  (with Nc/N = 7)
%  with $N_c=5$ and $\gamma=0.005$
\label{eq:superpx_illustrative_ex}
\end{figure}

We choose the transformation $\bW$ to be an (over)-segmentation of the image. More precisely, $\bY\bW$ computes an (over)-segmentation of the image $\bY$, and returns the average of all pixels inside each segmented region or superpixel. Note that the resulting pixels do not lie on a uniform sampling grid.
The conjugate transform, $\bY_{\!\mathcal{C}}\bW^{\ast}$, takes each segment in $\bY_{\!\mathcal{C}}$ and attributes its value to all pixels of the uniform image sampling grid that lie inside the corresponding region.
The successive application of both transforms, $\bW\bW^{\ast}$ effectively consists in averaging all pixels inside each segment of the input image. The decomposition of the Cuprite image using a segmentation and an over-segmentation algorithm is illustrated in Fig.~\ref{eq:superpx_illustrative_ex}.

\vspace{-1ex}
\section{Results}
\label{sec:results}

\cblue{
We \cblue{compare the performances} of the proposed MUA, the Total Variation (SUnSAL-TV), the \cblue{spatially} unregularized (SUnSAL) and the S$^2$WSU algorithms~\cite{bioucas2010SUnSAL,iordache2012sparseUnmixingTV,zhang2018S2WSU_sparseUnmixingReg}, both in terms of reconstruction error and computational complexity. 
The selection of these algorithms comes naturally since MUA, SUnSAL and SUnSAL-TV share the same sparse regression formulation, and  S$^2$WSU is considered a state-of-the-art algorithm for library-based sparse unmixing.}
For the proposed method, we compare two choices for the transformation $\bW$: 1) a binary partition tree based segmentation algorithm (BPT)~\cite{veganzones2014hyperspectralSegmentationBPT}, and 2) the simple linear iterative clustering (SLIC) over-segmentation method~\cite{achanta2012slicPAMI}. 
% Finally, we consider also the solution using the K-means algorithm, which does not take spatial information into account.
Finally, we consider also the solution using the K-means algorithm, which is not effective at taking local spatial information into account\footnote{\cblue{SLIC and K-means were implemented using the Euclidean distance between reflectance vectors (HI pixels).}}. 
%\cred{We refer to the MUA algorithm with each of these transformations as MUA-B, MUA-S, and MUA-K.}
% \cblue{For all three clustering-based methods we selected the cluster size among the integer values $\sqrt{N/K}\in\{3,\ldots,15\}$.}%, where $K$ is the number of clusters and $N$ is the number of pixel in the image.}

We considered a synthetic library $\bA_1\in\mathbb{R}^{224\times 240}$ generated by selecting a subset of 240 materials from the USGS library such that the angle between any pair of spectral signatures was at least $4.44$ degrees.

\vspace{-1ex}
\begin{table} [ht]
% \tiny
% \small
% \scriptsize
\footnotesize
%\caption{SRE results for unmixing data cubes DC1 and DC2 using SUnSAL, SUnSAL-TV, K-means, and MUA with BPT and SLIC.}
\caption{SRE results for unmixing data cubes DC1 and DC2.}
\vspace{-0.2cm}
\centering
\renewcommand{\arraystretch}{1.4}
\setlength\tabcolsep{3pt}
\resizebox{\linewidth}{!}{%
\begin{tabular}{c||c|c|c|c|c|c}
\hline\hline
\multicolumn{7}{c}{DC1 data cube}\\
\hline\hline
SNR & SUnSAL & SUnSAL-TV & \cblue{S$^2$WSU} & MUA$_{\text{{K-means}}}$ & MUA$_{\text{{BPT}}}$ & MUA$_{\text{{SLIC}}}$ \\
\hline\hline
20\,dB & 4.54\,dB & 9.42\,dB & 7.70\,dB & 9.96\,dB & \textbf{13.39}\,dB & 11.35\,dB \\
\hline
30\,dB & 8.91\,dB & 14.44\,dB & 15.49\,dB & 14.02\,dB & \textbf{18.26}\,dB & 15.73\,dB \\
\hline\hline
\multicolumn{7}{c}{DC2 data cube}\\
\hline\hline
SNR & SUnSAL & SUnSAL-TV & \cblue{S$^2$WSU} & MUA$_{\text{{K-means}}}$ & MUA$_{\text{{BPT}}}$ & MUA$_{\text{{SLIC}}}$ \\
\hline\hline
20\,dB & 4.27\,dB & 11.61\,dB & 9.39\,dB & 12.69\,dB & 14.08\,dB & \textbf{14.88}\,dB \\
\hline
30\,dB & 10.48\,dB & 17.97\,dB & \textbf{21.72}\,dB & 17.42\,dB & 16.92\,dB &  18.46\,dB \\
\hline\hline
\end{tabular}}
\label{tab:alg_param_dc1_dc2}
\end{table}

\vspace{-1ex}
\subsection{Simulation results using synthetic data sets}

For the simulations presented in this section two spatially correlated synthetic data cubes DC1 and DC2 were built using~5 and~9 endmembers, respectively, selected from library $\bA_1$. 
DC1 has 75$\times$75 pixels and its abundance map is composed of square regions distributed uniformly over a background in five rows. 
Data cube DC2 has 100$\times$100 pixels and its abundance maps were sampled according to a Dirichlet distribution centered at a Gaussian random field, leading to piecewise smooth maps that also have steep transitions.
For both datacubes, the generated HIs were contaminated by white Gaussian noise, with signal-to-noise ratios (SNR) of 20 and 30dB.
The quality of the reconstruction of the spectral mixtures was evaluated using the signal to reconstruction error, defined as $\text{SRE} =10\log_{10} ({\mathbb{E}\{\|\bX\|_F^2\}}/{\mathbb{E}\{\|\bX-\widehat{\!\bX}\|_F^2\}} )$ ~\cite{iordache2012sparseUnmixingTV}.
% \begin{equation} \label{eq:sre_metric}
% 	\text{SRE} = 10 \log_{10} \bigg(\frac{\mathbb{E}\{\|\bX\|_F^2\}}{\mathbb{E}\{\|\bX-\widehat{\!\bX}\|_F^2\}} \bigg)
%     \,\text{.}
% \end{equation}

\begin{figure}[htbp]
\centering
\includegraphics[width=0.485\textwidth]{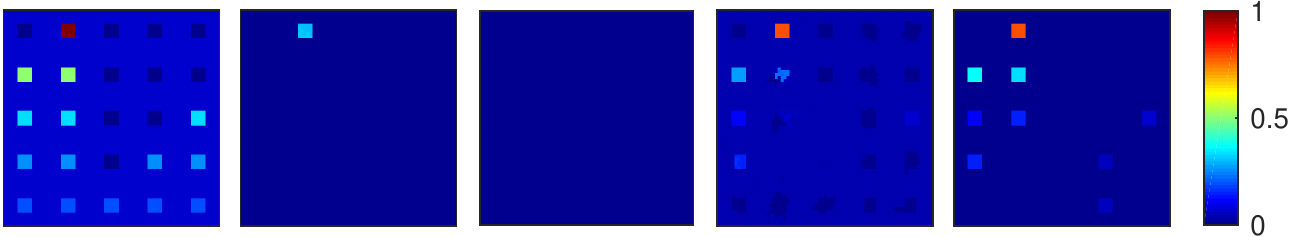} \\
\includegraphics[width=0.485\textwidth]{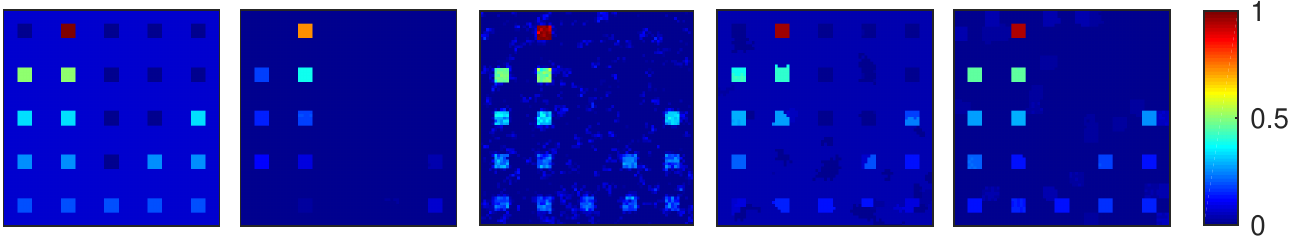} 
\begin{flushleft}
\vspace{-2ex}
{\footnotesize \hspace{3.5ex} True \hspace{3.2ex} SUnSAL-TV \hspace{1.7ex} S$^2$WSU \hspace{1.6ex} MUA (BPT) \hspace{0.2ex} MUA (SLIC)} \end{flushleft}
\vspace{-0.3cm}
\caption{Abundance maps estimated by the different unmixing methods for the 2nd endmember of data cube DC1. From top to bottom: SNR of 20 and 30dB.}
\label{fig:ex1_abundances_est}
\end{figure}

\begin{figure}[htbp]
\centering
\includegraphics[width=0.485\textwidth]{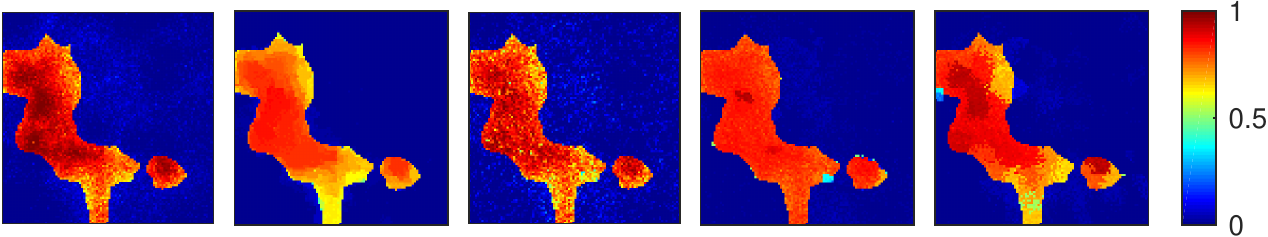} \\
\includegraphics[width=0.485\textwidth]{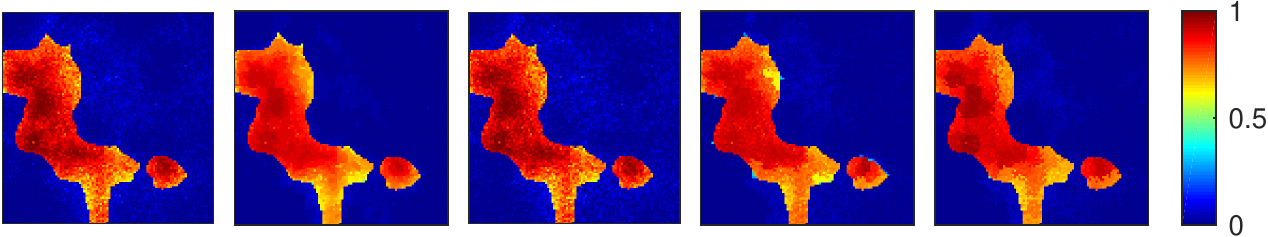} 
\begin{flushleft}
\vspace{-2ex}
{\footnotesize \hspace{3.5ex} True \hspace{3.2ex} SUnSAL-TV \hspace{1.7ex} S$^2$WSU \hspace{1.6ex} MUA (BPT) \hspace{0.2ex} MUA (SLIC)} \end{flushleft}
\vspace{-0.3cm}
\caption{Abundance maps estimated by the different unmixing methods for the 1st endmember of data cube DC2. From top to bottom: SNR of 20 and 30dB.}
\label{fig:ex2_abundances_est}
\end{figure}

To find the optimal parameters for the selected algorithms we performed a grid search for each dataset, and  the parameters leading to the best SRE results for each method were selected. For the MUA method, the parameter search occurred in the intervals $\lambda_{\cal{C}}\in[0.0001,\, 0.05]$, $\lambda\in[0.001,\, 0.1]$ and $\beta\in[0.007,\, 30]$, \cblue{while the cluster sizes were selected among the integer values $\sqrt{N/K}\in\{3,\ldots,15\}$.}
% \todo[inline]{The parameters for each MUA method (\cite{veganzones2014hyperspectralSegmentationBPT} and ~\cite{achanta2012slicPAMI}) are different, right? This should be better explained.}
For the \cblue{SUnSAL, SUnSAL-TV and S$^2$WSU} algorithms, the parameter ranges were selected according to those reported in the original work in~\cite{iordache2012sparseUnmixingTV}. \cblue{Due to space limitations, the selected parameters are available at~\cite{Borsoi_superpix_2017_arxiv}.}
The SRE achieved by the SUnSAL, SUnSAL-TV, \cblue{S$^2$WSU}, K-means and MUA are shown in Table~\ref{tab:alg_param_dc1_dc2} for both SNR values.
Samples of the reconstructed abundance maps for both data cubes and SNRs are shown in Figs.~\ref{fig:ex1_abundances_est} and~\ref{fig:ex2_abundances_est} for a qualitative comparison.

The computational complexity of the algorithms was evaluated through their execution times. SUnSAL, SUnSAL-TV, \cblue{S$^2$WSU}, BPT and SLIC were implemented using the codes made available by the authors. The algorithms were implemented in Matlab on a desktop computer equipped with an Intel Core I7 processor with 4.2Ghz and 16Gb RAM. The results are shown in Table~\ref{tab:alg_exec_time}.

% \begin{table}[!htbp]
% \small
% \caption{Execution time (in seconds) of the unmixing algorithms, averaged for all SNR values considered}
% \begin{center}
% \renewcommand{\arraystretch}{1.2}
% \begin{tabular}{c||c|c|c}
% \hline\hline
%  & SUnSAL & SUnSAL-TV & SUnSAL-M \\
% \hline\hline
% DC0 & 3.39\,s & 81.86\,s & 3.24\,s \\
% \hline
% DC1 & 2.66\,s & 79.16\,s & 2.53\,s \\
% \hline
% DC2 & 5.78\,s & 130.37\,s & 4.95\,s \\
% \hline
% Real Image & 184.8\,s & 1145.8\,s & 101.5\,s \\
% \hline
% \end{tabular}
% \end{center}
% \label{tab:alg_exec_time}
% \end{table}
\begin{table}[htbp]
% \small
% \scriptsize
\footnotesize
\caption{\cblue{Average Execution time (in seconds) of each algorithm}}
% \caption{Execution time (in seconds) of all unmixing algorithms, averaged for all SNR values considered.}
\vspace{-0.2cm}
\centering
\renewcommand{\arraystretch}{1.2}
\setlength\tabcolsep{3pt}
\resizebox{\linewidth}{!}{%
\begin{tabular}{c||c|c|c|c|c|c}
\hline\hline
 & SUnSAL & SUnSAL-TV & \cblue{S$^2$WSU} & MUA$_{\text{{K-means}}}$ & MUA$_{\text{{BPT}}}$ & MUA$_{\text{{SLIC}}}$ \\
\hline\hline
DC1 & 2.57\,s & 58.24\,s & 24.21\,s & 2.88\,s & 4.19\,s & 2.66\,s \\ 
\hline
DC2 & 4.24\,s & 92.1\,s & 42.41\,s & 3.69\,s & 4.94\,s & 4.04\,s \\ 
\hline
\makecell{Real \\[-0.1cm] Image} & 184.8\,s & 1145.8\,s & 469.5\,s & 84.9\,s & 77.1\,s & 101.5\,s \\
\hline
\end{tabular}}
\label{tab:alg_exec_time}
\end{table}

% \todo[inline]{I cannot notice any difference between the true images for SNRs of 20dB and 30dB!}

% \todo[inline]{Table~I indicates better results uisng MUA (BPT). However, the images seem to indicate that MUA (SLIC) leads to better results.  This is especially true for a 30dB SNR!}

% ----------------------------------------------------------
\subsubsection{Discussion}

It can be seen from Table~\ref{tab:alg_param_dc1_dc2} that the proposed algorithm can provide significantly better performance than the SUnSAL-TV algorithm for both data cubes.
The BPT segmentation-based transformation provided a variable performance, yielding very good results for DC1, but a performance closer to SUnSAL-TV for DC2, especially for SNR=30dB. This indicates a considerable sensitivity to the image content. The results obtained using SLIC, on the other hand, indicate a more regular performance, with significantly better results than SunSAL-TV for both data cubes.
\cblue{Although the $\text{S}^2$WSU presented the best SRE result for DC2 with SNR of 30dB, the method is very sensitive to variations of the noise level, as can be seen for both datasets.}
Finally, we note that a regularization based on the K-means algorithm performed only similarly to the SUnSAL-TV method, and significantly worse than the proposed transformations.
%\todo[inline]{But surprisingly good results for a method that completely disregards spatial information!}

Figs.~\ref{fig:ex1_abundances_est} and~\ref{fig:ex2_abundances_est} show samples of the abundance maps of data cubes DC1 and DC2 estimated by the SUnSAL-TV\cblue{, $\text{S}^2$WSU} and MUA algorithms using BPT and SLIC transformation, which provided the best quantitative performance\cblue{, except for the DC2 at 30dB where the $\text{S}^2$WSU produced a comparable map. However, the performance degradation of the $\text{S}^2$WSU is clear when the SNR is decreased}. 
% \todo[inline]{Can we state that K-means does not have results perceptually as good as these ones? From Table~I they should look better than those obtained using SunSAL-TV. Shouldn't we show the pictures?}
%
The results of the MUA algorithm were significantly better than those of the SUnSAL-TV algorithm. This difference is most noticeable for an SNR of 20dB, where the resulting abundance maps are much closer to the ground truth than those estimated by the SUnSAL-TV.% algorithm.
%
% The results of the MUA algorithm were significantly better than those of the SUnSAL-TV algorithm, specially for the over-segmentation-based transform, which showed a better preservation of object boundaries. This difference is most noticeable for an SNR of 20dB, where the resulting abundance maps are much closer to the ground truth than those estimated by the SUnSAL-TV.% algorithm.

% \todo[inline]{Really?! This is not very clear to me.}

% In terms of computational cost, MUA performed significantly better than SUnSAL-TV, with execution times comparable to those of SUnSAL algorithm, and on average~19 times smaller than those of SUnSAL-TV.
%
In terms of computational cost, MUA performed significantly better than SUnSAL-TV, with execution times comparable to those of SUnSAL algorithm, \cblue{and, on average,~19 and 10 times smaller than those of SUnSAL-TV and $\text{S}^2$WSU respectively.}
These results illustrate the effectiveness of the proposed regularization method both in terms of quality and computational cost.

\begin{figure}[htbp]
\centering
\includegraphics[width=0.485\textwidth]{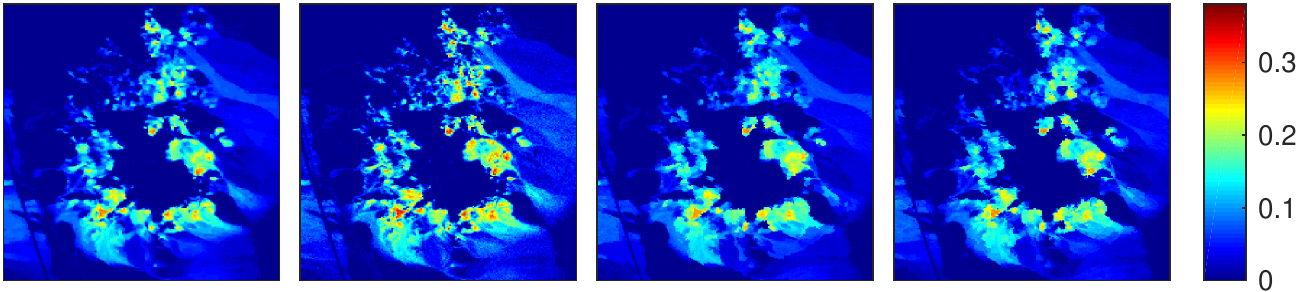} \\
\includegraphics[width=0.485\textwidth]{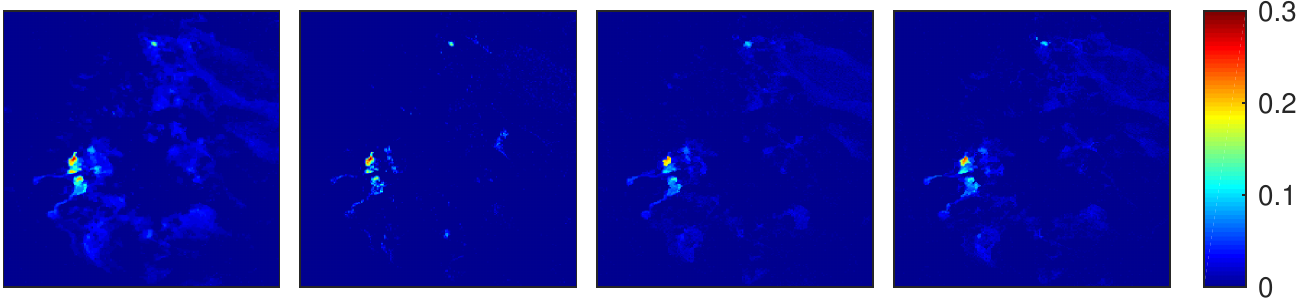} 
\\
\includegraphics[width=0.485\textwidth]{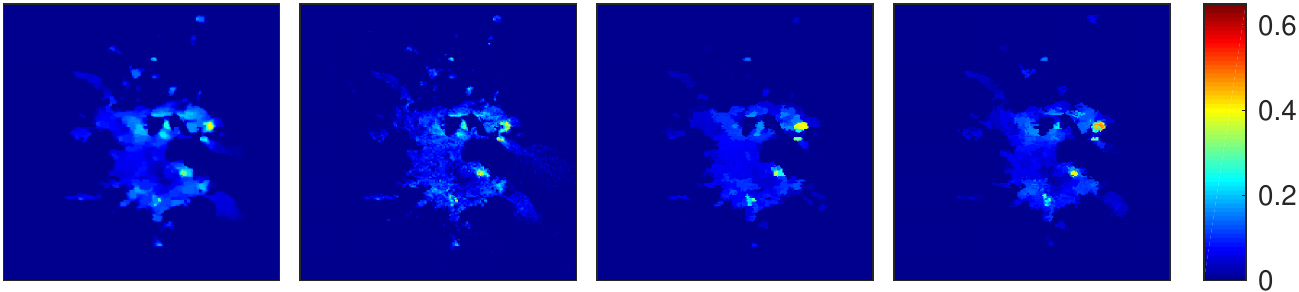} 
\begin{flushleft}
\vspace{-2ex}
{\footnotesize \hspace{1.3ex} SUnSAL-TV \hspace{4.7ex} S$^2$WSU \hspace{4.7ex} MUA (BPT) \hspace{2.7ex} MUA (SLIC)} \end{flushleft}
\vspace{-0.3cm}
\caption{Fractional abundance maps estimated for the Cuprite image. From top to bottom: Alunite, Buddingtonite, and Chalcedony.}% From left to right: SUnSAL-TV, \cred{S$^2$WSU}, MUA (BPT) and MUA (SLIC).}
\label{fig:real_abundances_est}
\end{figure}

\vspace{-2ex}
% ----------------------------------------------------------
\subsection{Simulation results using real image}

In this experiment, we consider a well-known region of the Cuprite data set with 250$\times$191 pixels. The spectral library $\bA\in\mathbb{R}^{188\times498}$ was built using the spectral signatures in the USGS library after removing water absorption and low SNR spectral bands, resulting in 188 bands.
%Spectral bands presenting water absorption and low SNR were removed from the image, resulting in 188 spectral bands.
%
%The spectral signatures from the USGS library with the corresponding bands removed were used as the spectral library $\bA\in\mathbb{R}^{188\times498}$.
%
% In order to minimize calibration mismatches between the real pixel spectra and the elements of the spectral library, the band-dependent correction strategy described in~\cite{iordache2011sparseUnmixing} was applied to the image.
%
% The parameters of the algorithms were selected empirically for MUA, and set identically to those reported in~\cite{iordache2012sparseUnmixingTV} for SUnSAL and SUnSAL-TV\cblue{, and identically to those in~\cite{zhang2018S2WSU_sparseUnmixingReg} for the $\text{S}^2$WSU}.
%
The parameters of the algorithms were selected empirically for MUA, and set identically to those reported in~\cite{iordache2012sparseUnmixingTV} for SUnSAL and SUnSAL-TV\cblue{, and for the $\text{S}^2$WSU we used $\lambda_{\text{swsp}} = 7\times 10^{-5}$}.
Since the true abundance maps are unavailable for this HI, we compare the fractional abundance maps of three dominant materials (Alunite, Buddingtonite, and Chalcedony) estimated using the three algorithms. The results are shown in Fig.~\ref{fig:real_abundances_est}.

%we make a qualitative assessment of the recovered abundance maps based on reference maps of minerals known to be present in prominent fashion in the Cuprite mining district. %\footnote{http://speclab.cr.usgs.gov/cuprite95.tgif.2.2um\_map.gif}.
%
%A qualitative comparison of the fractional abundance maps of three dominant materials (Alunite, Buddingtonite, and Chalcedony) estimated using the three algorithms is shown in Fig.~\ref{fig:real_abundances_est}. 

% Although all algorithms estimated considerable abundances for the materials known to be present in the scene, the use of spatial regularization resulted in more spatially consistent estimates, with less outliers resulting from the influence of measurement noise.
% %
% This can be observed in the Buddingtonite abundance maps, where the SUnSAL results contain regions only sparsely filled with non-zero abundances, whereas SUnSAL-TV and SUnSAL-M provided more homogeneous estimates of the abundances in those regions where the material is present.

Although the unmixing results for SUnSAL-TV\cblue{, $\text{S}^2$WSU,} and MUA were similar, it can be observed that the TV regularization tends to yield an over-smooth visual effect.
This is not observed in the results using \cblue{$\text{S}^2$WSU and  MUA (especially for the over-segmentation transformation) which produce} spatially consistent abundance maps without compromising the fine variability and the intricate structures in the image.
% This is not observed in the results using \cblue{$\text{S}^2$WSU and}  MUA, especially for the over-segmentation transformation, which produces spatially consistent abundance maps without compromising the fine variability and the intricate structures in the image.
%
% \cblue{The $\text{S}^2$WSU on the other hand, }
%
These results again indicate the effectiveness of the proposed spatial regularization.
The computational times are shown in Table~\ref{tab:alg_exec_time}, and illustrate again the considerably lower complexity of MUA when compared to SUnSAL-TV \cblue{and $\text{S}^2$WSU}. It also runs significantly faster than the SUnSAL algorithm due to the faster convergence rate achieved with the use of proposed regularization.
\cblue{Further simulations with different real datasets are also available at~\cite{Borsoi_superpix_2017_arxiv}.}

% \begin{table} [!htbp]
% \small
% \caption{Parameters used in the Cuprite image simulations.}
% \begin{center}
% % \renewcommand{\arraystretch}{1.2}
% \begin{tabular}{c|c|c}
% \hline\hline
% SUnSAL & SUnSAL-TV & SUnSAL-M \\
% \hline\hline
% \multirow{2}{*}{$\lambda=10^{-3}$} & $\lambda=10^{-3}$ & $\lambda_{\mathcal{C}}=0.001$, $\lambda=0.001$\\
% & $\lambda_{\text{TV}}=10^{-3}$ & $\beta=3$ \\
% \hline
% \end{tabular}
% \end{center}
% \label{tab:alg_params_real_img}
% \end{table}

% ----------------------------------------------------------
\section{Conclusions}
\label{sec:conclusions}

% In this paper, we presented a novel multiscale methodology to couple with the spatial regularization of sparse HU problems. The presented approach allowed one to decompose the spatially regularized unmixing problem into two simple, low-cost problems in different domains. Although many domain transformation could be considered with the proposed strategy, the superpixel decomposition captured the spatial contextual information of the fractional abundances at reasonable computational cost. 
% %
% Simulation results with both synthetic and real data showed that the proposed method outperforms both,  SUnSAL and SUnSAL-TV, in almost all scenarios, with Execution Time comparable to the SUnSAL. In fact, the proposed method solved the problems 10 to 31 times faster then the SUnSAL-TV depending on factors such as  the number of pixels, the number of bands, and problem conditioning. Simulations also showed that the improvement in the problem conditioning, when using the proposed methodology, provided Execution Times that can be even smaller than the SUnSAL algorithm. 

In this paper, we presented a novel multiscale methodology to introduce spatial information in sparse HU problems. It decomposes the spatially regularized unmixing problem into two simple, low-cost problems in different image domains. 
Two multiscale domain transformations were proposed based on segmentation and over-segmentation methods, which allow an effective capture of spatial and spectral contextual informations at a reasonable computational cost. 
Simulation results using both synthetic and real data showed that the proposed method outperforms state-of-the-art TV-based sparse HU algorithms. Moreover, it requires execution times that are an order of magnitude lower than the TV-based solution, and comparable to or even smaller than those of unregularized methods.

\bibliographystyle{IEEEtran}
\bibliography{references}

\onecolumn
\centerline{{\huge Supplemental Material}}

\bigskip
The following material supplements the paper 

\bigskip
\noindent [A] R. A. Borsoi, T. Imbiriba, J. C. M. Bermudez and C. Richard, ``A Fast Multiscale Spatial Regularization for Sparse Hyperspectral Unmixing''. IEEE Geoscience and Remote Sensing Letters, 2018. DOI: 10.1109/LGRS.2018.2878394.

\setcounter{section}{0}
\setcounter{table}{0}
\setcounter{figure}{0}

\section{Parameters of the algorithms}
The parameters of the algorithms for all simulations are depicted in Tables~\ref{tab:alg_parameterss1} to~\ref{tab:alg_parameterss3}. Table~\ref{tab:alg_parameterss1} and Table~\ref{tab:alg_parameterss2} provide detailed information about the parameters used to obtain the results shown in Table~I of [A]. Table~\ref{tab:alg_parameterss1} shows the values of the regularization parameters, and  Table~\ref{tab:alg_parameterss2} shows the numbers of clusters generated by applying each of the three image segmentation methods in MUA. Table~\ref{tab:alg_parameterss3} provides the parameter values used for each of the four algorithms whose Culprite image unmixing performances are compared in Fig.~4 of [A].

\begin{table} [ht]
% \tiny
\small
% \scriptsize
% \footnotesize
%\caption{SRE results for unmixing data cubes DC1 and DC2 using SUnSAL, SUnSAL-TV, K-means, and MUA with BPT and SLIC.}
\caption{Parameters of each algorithm}
\vspace{-0.2cm}
\centering
\renewcommand{\arraystretch}{1.4}
\setlength\tabcolsep{5pt}
\begin{tabular}{c||c|c|c|c|c|c}
\hline\hline
\multicolumn{7}{c}{DC1 data cube}\\
\hline\hline
SNR  & SUnSAL & SUnSAL-TV & MUA$_{\text{{K-means}}}$ & MUA$_{\text{{BPT}}}$ & MUA$_{\text{{SLIC}}}$ & {S$^2$WSU} \\
\hline\hline
20\,dB & $\lambda=0.7$ & \makecell{$\lambda=0.05$\\$\lambda_{TV}=0.05$} &
\makecell{$\lambda_C=0.005$\\$\lambda=0.5$\\$\beta=30$} & 
\makecell{$\lambda_C=0.005$\\$\lambda=0.1$\\$\beta=30$} & 
\makecell{$\lambda_C=0.03$\\$\lambda=0.1$\\$\beta=30$} &
$\lambda=0.1$ \\
\hline
30\,dB & $\lambda=0.1$ & \makecell{$\lambda=0.007$\\$\lambda_{TV}=0.01$} &
\makecell{$\lambda_C=0.005$\\$\lambda=0.05$\\$\beta=10$} & 
\makecell{$\lambda_C=0.005$\\$\lambda=0.1$\\$\beta=30$} & 
\makecell{$\lambda_C=0.007$\\$\lambda=0.05$\\$\beta=10$} &
$\lambda=0.005$\\
\hline\hline
\multicolumn{7}{c}{DC2 data cube}\\
\hline\hline
SNR  & SUnSAL & SUnSAL-TV & MUA$_{\text{{K-means}}}$ & MUA$_{\text{{BPT}}}$ & MUA$_{\text{{SLIC}}}$ & {S$^2$WSU} \\
\hline\hline
20\,dB & $\lambda=0.1$ & \makecell{$\lambda=0.01$\\$\lambda_{TV}=0.03$} &
\makecell{$\lambda_C=0.005$\\$\lambda=0.5$\\$\beta=10$} & 
\makecell{$\lambda_C=0.005$\\$\lambda=0.1$\\$\beta=5$} & 
\makecell{$\lambda_C=0.007$\\$\lambda=0.1$\\$\beta=10$} &
$\lambda=0.01$ \\
\hline
30\,dB & $\lambda=0.01$ & \makecell{$\lambda=0.005$\\$\lambda_{TV}=0.007$} &
\makecell{$\lambda_C=0.005$\\$\lambda=0.01$\\$\beta=1$} & 
\makecell{$\lambda_C=0.001$\\$\lambda=0.05$\\$\beta=1$} & 
\makecell{$\lambda_C=0.003$\\$\lambda=0.03$\\$\beta=3$} &
$\lambda=0.01$ \\
%
%
% \hline\hline
% \multicolumn{7}{c}{Cuprite image}\\
% \hline\hline
%   & SUnSAL & SUnSAL-TV & MUA$_{\text{{K-means}}}$ & MUA$_{\text{{BPT}}}$ & MUA$_{\text{{SLIC}}}$ & {S$^2$WSU} \\
% \hline\hline
%  & $\lambda=$ & \makecell{$\lambda=$\\$\lambda_{TV}=$} &
% \makecell{$\lambda_C=$\\$\lambda=$\\$\beta=$} & 
% \makecell{$\lambda_C=$\\$\lambda=$\\$\beta=$} & 
% \makecell{$\lambda_C=$\\$\lambda=$\\$\beta=$} &
% $\lambda=$ \\
\hline\hline
\end{tabular}
\label{tab:alg_parameterss1}
\end{table}

\begin{table} [ht]
% \tiny
\small
% \scriptsize
% \footnotesize
%\caption{SRE results for unmixing data cubes DC1 and DC2 using SUnSAL, SUnSAL-TV, K-means, and MUA with BPT and SLIC.}
\caption{Parameters of each algorithm}
\vspace{-0.2cm}
\centering
\renewcommand{\arraystretch}{1.5}
\setlength\tabcolsep{5pt}
\begin{tabular}{c|c|c|c}
\hline\hline
\multicolumn{4}{c}{Cuprite image}\\
\hline\hline
SUnSAL-TV & MUA$_{\text{{BPT}}}$ & MUA$_{\text{{SLIC}}}$ & {S$^2$WSU} \\
\hline\hline
\makecell{$\lambda=0.001$\\$\lambda_{TV}=0.001$} &
\makecell{$\lambda_C=0.001$\\$\lambda=0.001$\\$\beta=3$} & 
\makecell{$\lambda_C=0.001$\\$\lambda=0.001$\\$\beta=3$} &
$\lambda=7\times10^{-5}$ \\
\hline\hline
\end{tabular}
\label{tab:alg_parameterss2}
\end{table}

\begin{table} [ht]
% \tiny
\small
% \scriptsize
% \footnotesize
%\caption{SRE results for unmixing data cubes DC1 and DC2 using SUnSAL, SUnSAL-TV, K-means, and MUA with BPT and SLIC.}
\caption{Number of clusters of the MUA algorithm}
\vspace{-0.2cm}
\centering
\renewcommand{\arraystretch}{1.4}
\setlength\tabcolsep{5pt}
\begin{tabular}{c||c|c|c}
\hline\hline
\multicolumn{4}{c}{DC1 data cube}\\
\hline\hline
SNR   & MUA$_{\text{{K-means}}}$ & MUA$_{\text{{BPT}}}$ & MUA$_{\text{{SLIC}}}$ \\
\hline\hline
20\,dB & $\sqrt{K/N}=13$ & $\sqrt{K/N}=14$ & $\sqrt{K/N}=6$ \\
\hline
30\,dB & $\sqrt{K/N}=7$ & $\sqrt{K/N}=10$ & $\sqrt{K/N}=5$ \\
\hline\hline
\multicolumn{4}{c}{DC2 data cube}\\
\hline\hline
SNR  & MUA$_{\text{{K-means}}}$ & MUA$_{\text{{BPT}}}$ & MUA$_{\text{{SLIC}}}$ \\
\hline\hline
20\,dB & $\sqrt{K/N}=11$ & $\sqrt{K/N}=13$ & $\sqrt{K/N}=8$ \\
\hline
30\,dB & $\sqrt{K/N}=8$ & $\sqrt{K/N}=11$ & $\sqrt{K/N}=7$ \\
\hline\hline
\multicolumn{4}{c}{Cuprite image}\\
\hline\hline
   & MUA$_{\text{{K-means}}}$ & MUA$_{\text{{BPT}}}$ & MUA$_{\text{{SLIC}}}$ \\
\hline\hline
 & $\times$ & $\sqrt{K/N}=7$ & $\sqrt{K/N}=5$ \\
\hline\hline
\end{tabular}
\label{tab:alg_parameterss1}
\end{table}

\section{Additional simulations with real data sets}

\begin{figure}[th]
\centering
\begin{minipage}[t]{.3\linewidth}
  \centering
%   \centerline{\includegraphics[width=\linewidth,trim={0cm 3cm 0cm 0},clip]{realimg/img_example.pdf}}
\centerline{\includegraphics[width=\linewidth]{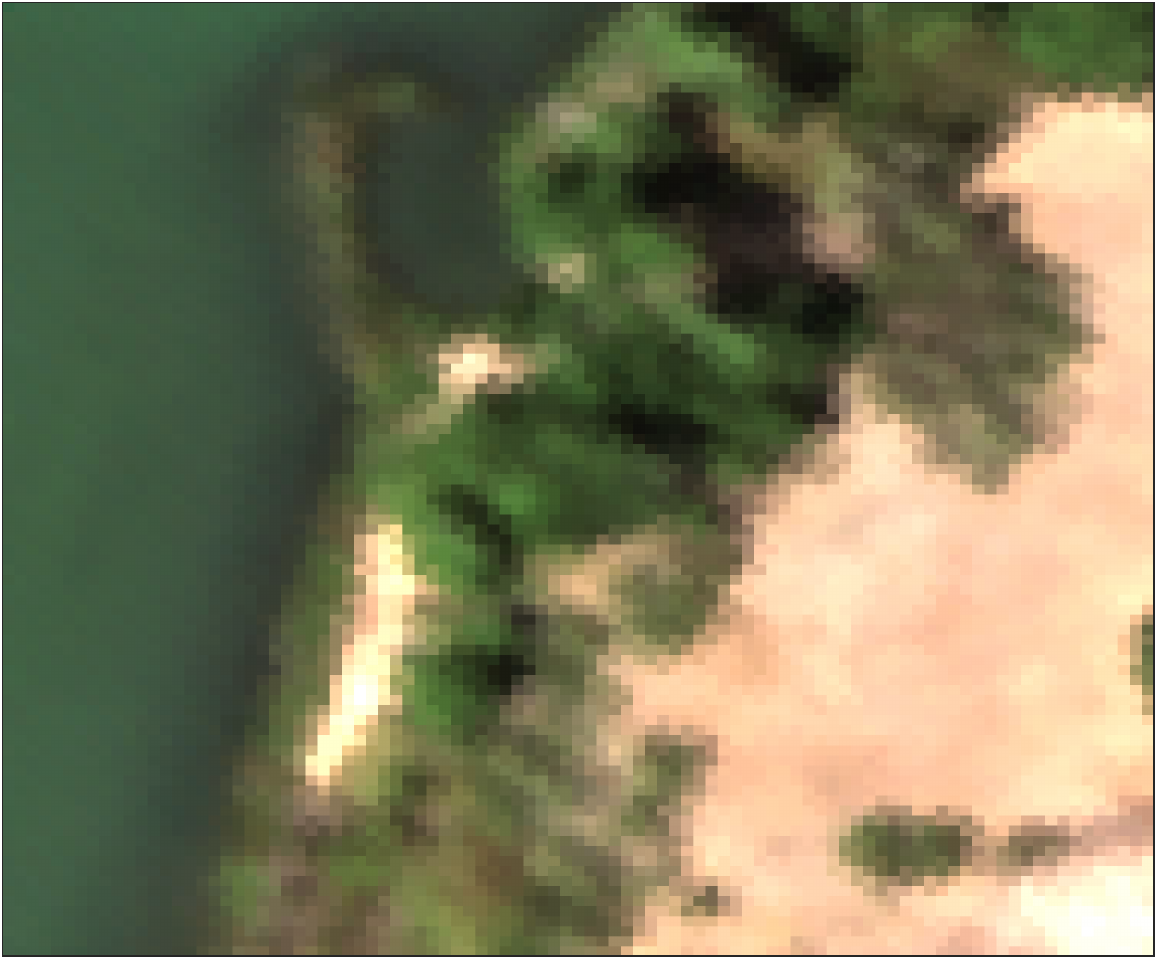}}
  \centerline{Samson}\medskip
\end{minipage}
% \hfill
\hspace{0.1cm}
\begin{minipage}[t]{0.3\linewidth}
  \centering
%   \centerline{\includegraphics[width=\linewidth,trim={0 3cm 0 0},clip]{realimg/img_example_BPT.pdf}}
\centerline{\includegraphics[width=\linewidth]{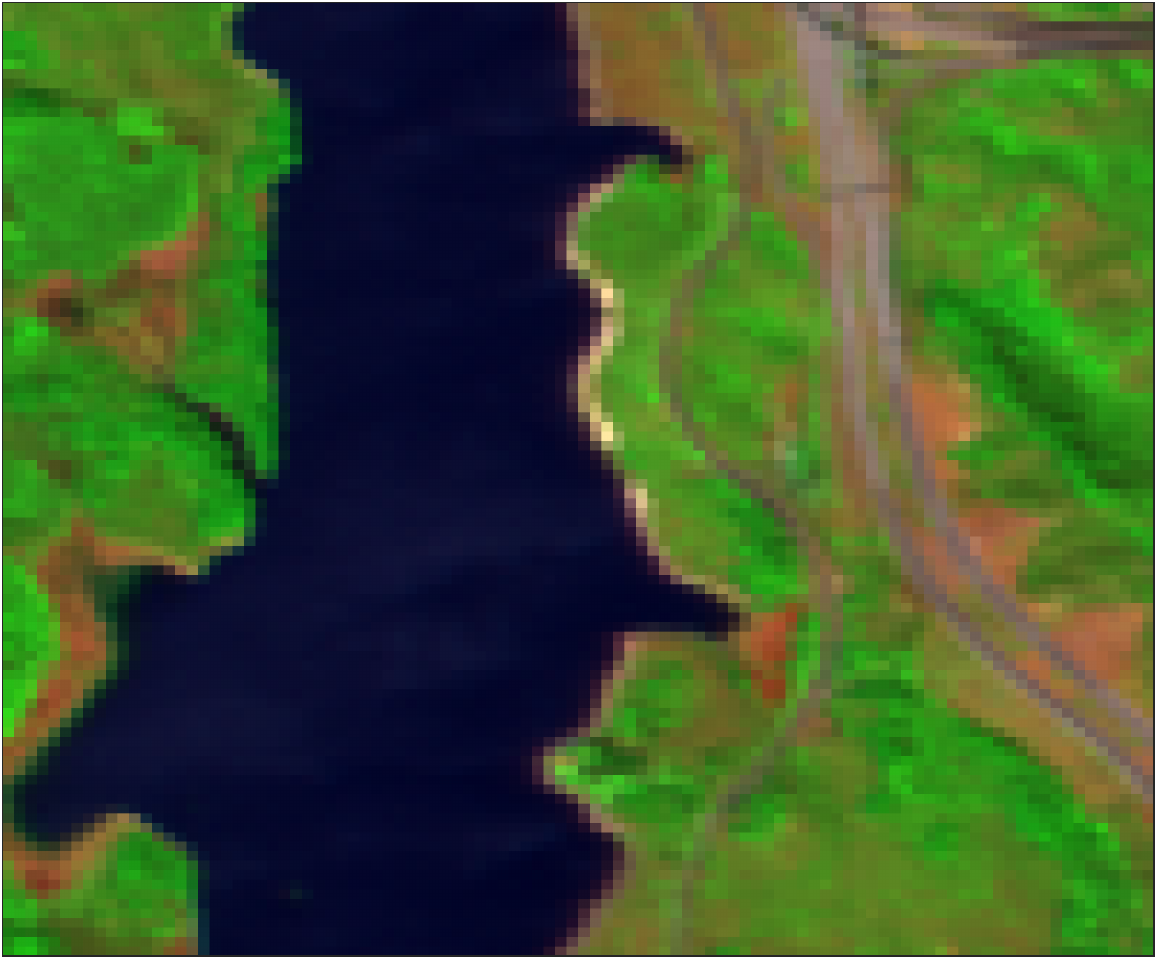}}
  \centerline{Jasper Ridge}\medskip
\end{minipage}
% \hfill
\hspace{0.1cm}
\begin{minipage}[t]{0.3\linewidth}
  \centering
%   \centerline{\includegraphics[width=\linewidth,trim={0 3cm 0 0},clip]{realimg/img_example_sppx.pdf}}
\centerline{\includegraphics[width=\linewidth]{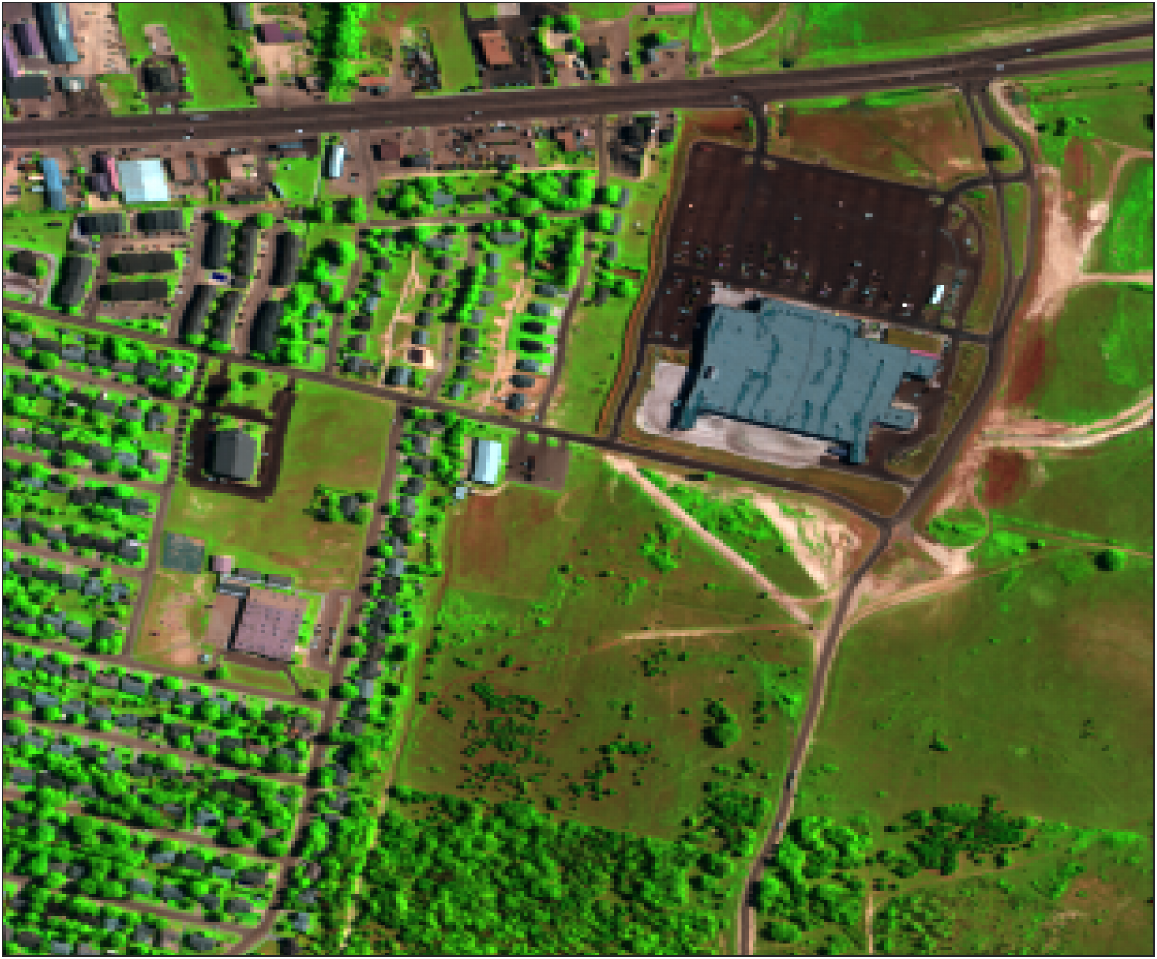}}
  \centerline{Urban}\medskip
\end{minipage}
\vspace{-0.2cm}
\caption{HIs used in the additional experiments.}
\label{fig:ims_additional}
\end{figure}

In addition to the results shown in [A] for the Cuprite data set, we consider here three more HIs to illustrate the performance of the proposed method, namely: the Samson, Jasper Ridge and Urban HIs, depicted in Fig.~\ref{fig:ims_additional}.

Since spectral libraries containing materials found in those scenes were not available \textit{a priori}, we employed an image-based library construction method to extract $\bA$ directly from the image~\cite{Somers12}. Specifically, we first constructed a preliminary library $\bA$ by manually extracting pure pixels from the observed images. Afterwards, the libraries for each material were pruned in order to limit the amount of signatures for each material, leading to libraries of size $P=105$ for the Samson HI, $P=529$ for the Jasper Ridge, and $P=651$ for the Urban HI.
%
% Since the estimated SNR of these images was very close to 30dB, we employed the same parameters as in DC2-30dB for all algorithms.
%
For the Samson and Jasper Ridge HIs, we used the same parameters as in DC2-30dB for all algorithms. For the Urban HI, the parameters of all algorithms were adjusted manually and are depicted in Table~\ref{tab:alg_parameterss3}.

Afterwards, we have summed the abundance contributions of all signatures of the same material and normalized the result such that the abundance sum-to-one constraint was satisfied. This allowed us to display a single set of abundance maps for each underlying material.
The results are depicted in Figs.~\ref{fig:abundances_samson},~\ref{fig:abundances_jasper}, and~\ref{fig:abundances_urban} where the abundance maps are depicted for each endmember (rows) and the SUnSAL-TV, S$^2$WSU, MUA(BPT) and MUA(SLIC) algorithms.

\subsection{Discussion}
\begin{figure}[th]
\centerline{\includegraphics[width=0.8\linewidth]{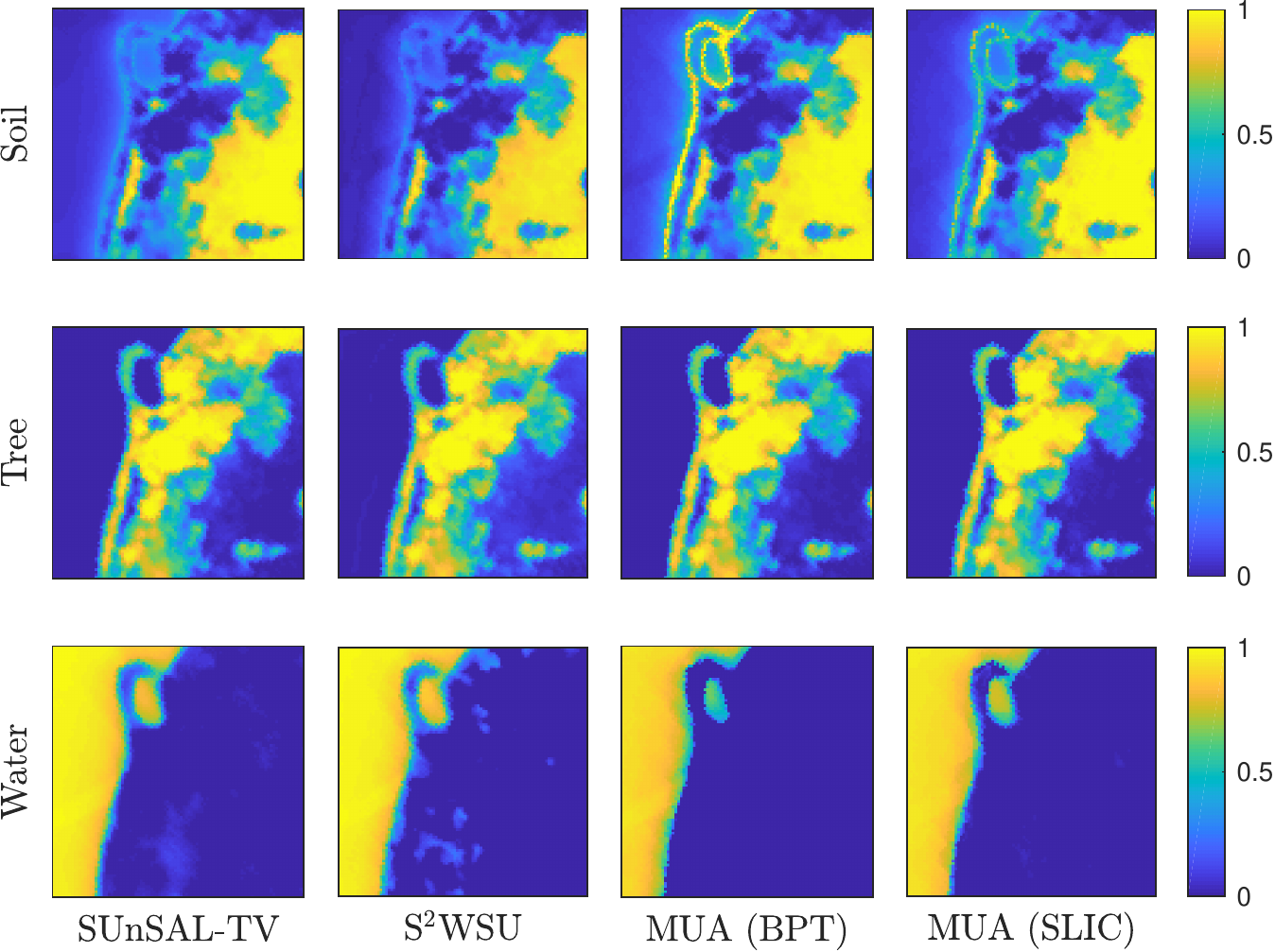}}
\caption{Reconstructed abundance maps for the Samson HI.}
\label{fig:abundances_samson}
\end{figure}

\begin{figure}[th]
\centerline{\includegraphics[width=0.85\linewidth]{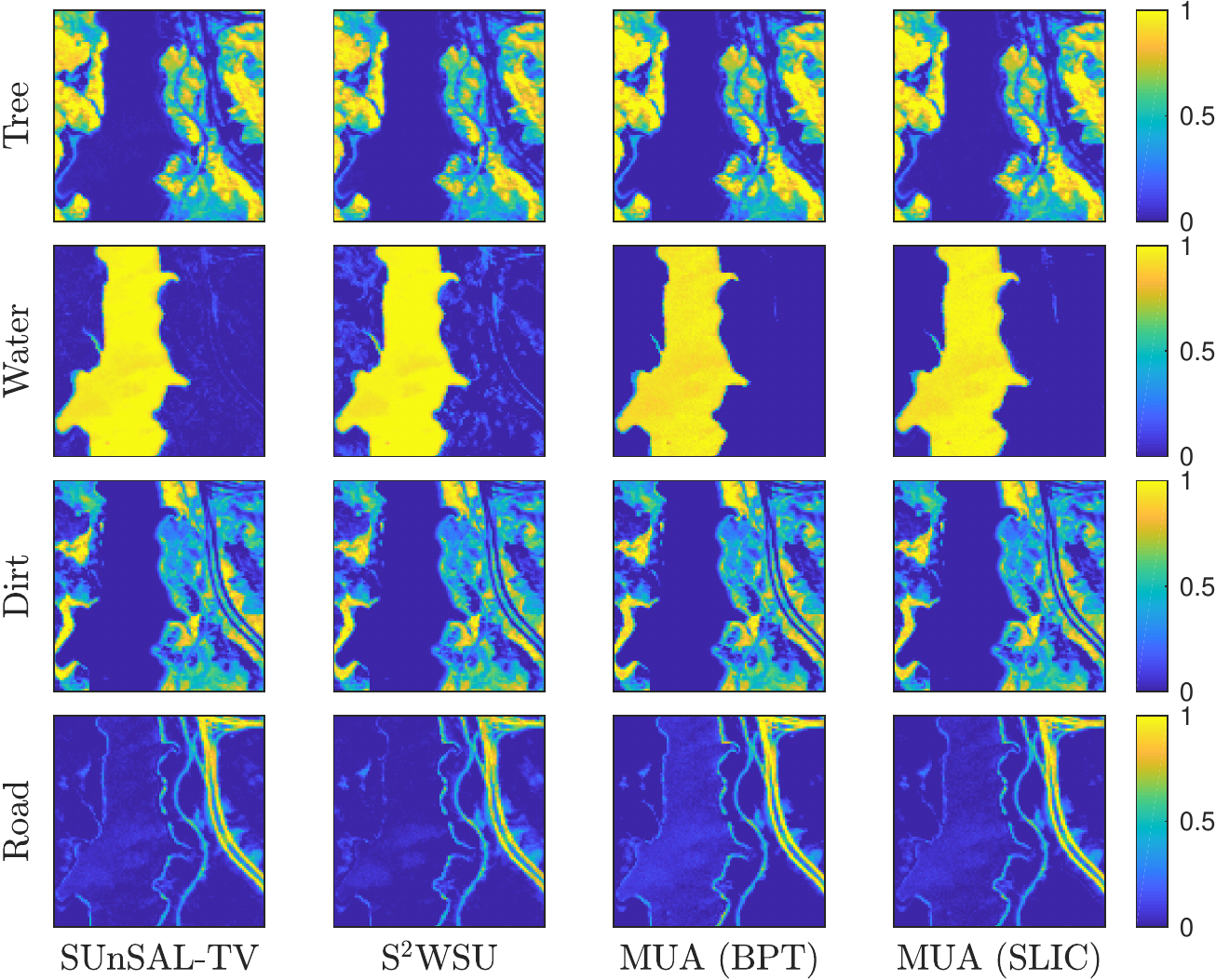}}
\caption{Reconstructed abundance maps for the Jasper Ridge HI.}
\label{fig:abundances_jasper}
\end{figure}

\begin{figure}[th]
% \centerline{\includegraphics[width=0.8\linewidth]{newIms/estim_abundances_urban-crop}}
\centerline{\includegraphics[width=0.85\linewidth]{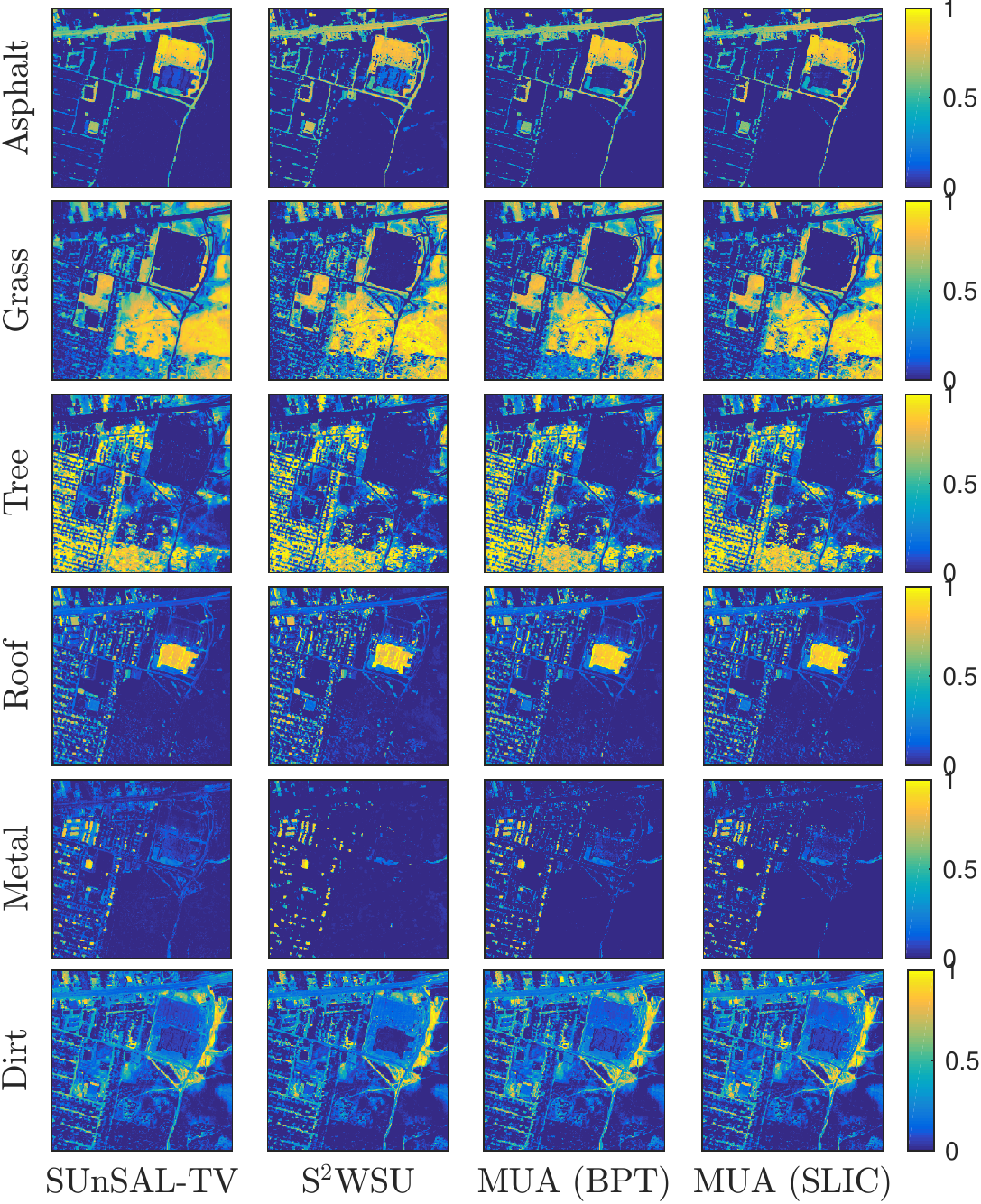}}
\caption{Reconstructed abundance maps for the Urban HI.}
\label{fig:abundances_urban}
\end{figure}

Fig.~\ref{fig:abundances_samson} shows that the MUA algorithm using either segmentation (BPT) and over-segmentation (SLIC) methods leads to more accurate and smooth abundance maps. The improvement is more evident in the Soil and Water abundance maps, in which the proposed strategy provided more uniform maps than the ones obtained with the competing algorithms. The same behavior can noticed in Figs.~\ref{fig:abundances_jasper} and~\ref{fig:abundances_urban}. Fig.~\ref{fig:abundances_jasper} shows a clear improvement in the water abundance map when comparing the proposed method with the competing algorithms. Finally, the last simulation also shows that the proposed strategies yielded  smooth abundance maps, which are comparable with those obtained using the competing algorithms and have a good compromise between smoothness and detail. This illustrates that the proposed strategy can provide abundance estimations that are at least as good as state of the art methods.

% \cblue{Thus, the simulations show that the proposed strategy can provide abundance estimations that are at least as good as the state of the art methods with a fraction of its computational complexity.}

\begin{table} [ht]
% \tiny
\small
% \scriptsize
% \footnotesize
%\caption{SRE results for unmixing data cubes DC1 and DC2 using SUnSAL, SUnSAL-TV, K-means, and MUA with BPT and SLIC.}
\caption{Parameters of each algorithm}
\vspace{-0.2cm}
\centering
\renewcommand{\arraystretch}{1.5}
\setlength\tabcolsep{5pt}
\begin{tabular}{c|c|c|c}
\hline\hline
\multicolumn{4}{c}{Urban image}\\
\hline\hline
SUnSAL-TV & MUA$_{\text{{BPT}}}$ & MUA$_{\text{{SLIC}}}$ & {S$^2$WSU} \\
\hline\hline
\makecell{$\lambda=0.005$\\$\lambda_{TV}=0.007$} &
\makecell{$\lambda_C=0.005$\\$\lambda=0.01$\\$\beta=3$} & 
\makecell{$\lambda_C=0.005$\\$\lambda=0.01$\\$\beta=3$} &
$\lambda=0.02$ \\
\hline\hline
\end{tabular}
\label{tab:alg_parameterss3}
\end{table}
% BPT
% lambda1_sp = 0.005; 0.01;
% lambda2_sp = 0.01;
% beta       = 3;
% sideBPT    = 7;
% SLIC
% lambda1_sp = 0.005; 0.01;
% lambda2_sp = 0.01;
% beta       = 3;
% slic_size  = 5;
% slic_reg   = 0.0025;
% TV
% lambda    = 0.005;
% lambda_TV = 0.007;
% S2WSU
% lambda_swsp = 20e-3; 

\end{document}